# LightLDA: Big Topic Models on Modest Compute Clusters

Jinhui Yuan[1], Fei Gao[1,2], Qirong Ho[3], Wei Dai[4], Jinliang Wei[4], Xun Zheng[4],

Eric P. Xing[4], Tie-Yan Liu[1], Wei-Ying Ma[1]

**Abstract**

When building large-scale machine learning (ML) programs, such as massive topics models or deep networks with up to trillions of parameters and training examples, one usually assumes that such massive tasks can only be attempted with industrial-sized clusters with thousands of nodes, which are out of reach for most practitioners or academic researchers. We consider this challenge in the context of topic modeling on web-scale corpora, and show that with a modest cluster of as few as 8 machines, we can train a topic model with 1 million topics and a 1-million-word vocabulary (for a total of 1 trillion parameters), on a document collection with 200 billion tokens — a scale not yet reported even with thousands of machines. Our major contributions include: 1) a new, highly-efficient $\mathcal{O}(1)$ Metropolis-Hastings sampling algorithm, whose running cost is (surprisingly) agnostic of model size, and empirically converges nearly an order of magnitude more quickly than current state-of-the-art Gibbs samplers; 2) a structure-aware model-parallel scheme, which leverages dependencies within the topic model, yielding a sampling strategy that is frugal on machine memory and network communication; 3) a differential data-structure for model storage, which uses separate data structures for high- and low-frequency words to allow extremely large models to fit in memory, while maintaining high inference speed; and 4) a bounded asynchronous data-parallel scheme, which allows efficient distributed processing of massive data via a parameter server. Our distribution strategy is an instance of the model-and-data-parallel programming model underlying the Petuum framework for general distributed ML, and was implemented on top of the Petuum open-source system. We provide experimental evidence showing how this development puts massive models within reach on a small cluster while still enjoying proportional time cost reductions with increasing cluster size, in comparison with alternative options.

## 1 Introduction

Topic models (TM) are a popular and important modern machine learning technology that has been widely used in text mining, network analysis and genetics, and more other domains [17, 4, 23, 25, 2]. Their impact on the technology sector and the Internet has been tremendous — numerous companies having developed their large-scale TM implementations [12, 1, 21], with applications to advertising and recommender systems. A key goal of contemporary research is to scale TMs, particularly the Latent Dirichlet Allocation (LDA) model [4], to web-scale corpora (Big Data). Crucially, internet-scale corpora are significantly more complex than smaller, well-curated document collections, and thus require high-capacity topic parameter spaces featuring up to millions of topics and vocabulary words (and hence trillions of parameters, i.e. Big Models), in order to capture long-tail semantic information that would otherwise be lost when learning only a few thousand topics [21].

To achieve such massive data and model scales, one approach would be to engineer a distributed system that can efficiently execute well-established *data-parallel* strategies (i.e., split documents over workers which have shared access to all topic parameters) for LDA [15], while applying algorithmic speedups such as the SparseLDA [22] or AliasLDA [10] samplers to further decrease running times. Such efforts have enabled LDA models with 10s of billions of parameters to be inferred from billions of documents, using up to thousands of machines [12, 1, 11, 21]. While such achievements are impressive, they are unfortunately costly to run: for instance, a cluster of 1000 machines will cost millions of dollars to set up (not to mention the future costs of power and regular maintenance). Alternatively, one might rent equivalent compute capacity from a cloud provider, but given the current typical price of $\geq \$1$ per hour per research-grade machine, a single month of operations would cost $\geq \$700,000$. Neither option is feasible for the majority of researchers and practitioners, who are likely to have more modest budgets.

---

[1] Microsoft Research. Email: {jiyuan,tie-yan.liu,wyma}@microsoft.com
[2] Department of Computer Science & Engineering, Beihang University. Email: gf0109@gmail.com
[3] Institute for Infocomm Research, A*STAR, Singapore. Email: hoqirong@gmail.com
[4] School of Computer Science, Carnegie Mellon University. Email: {wdai,jinliangw,xunzheng,epxing}@cs.cmu.edu

Rather than insisting that big computing is the only way to solve large ML problems, what if we could bring large topic models with trillions of parameters to a more cost-effective level, by providing an implementation that is efficient enough for modest clusters with at most 10s of machines? We approach this problem at three levels: (1) we distribute LDA inference in a *data- and model-parallel* fashion: both data and model are partitioned and then streamed across machines, so as to make efficient use of memory and network resources within a distributed cluster; (2) we develop an Metropolis-Hastings sampler with carefully constructed proposals that allows for $\mathcal{O}(1)$ amortized sampling time per word/token, resulting in a high convergence rate (in terms of real time) that beats existing state-of-the-art samplers by a significant margin [22, 10]; (3) we employ a differential data structure to leverage the fact that that web-scale corpora exhibit both high-frequency power-law words as well as low-frequency long-tail words, which can be treated differently in storage, resulting in high memory efficiency without a performance penalty from a homogeneous data structure.

By realizing these ideas using the open-source Petuum framework (www.petuum.org) [8, 9, 5], we have produced a compute-and-memory efficient distributed LDA implementation, LightLDA, that can learn an LDA model with 1 trillion model parameters (1m topics by 1m vocabulary words) from billions of documents (200 billion tokens), on a compute cluster with as few as 8 standard machines (whose configuration is roughly similar to a typical compute instance from a cloud provider) in 180 hours, which proportionally drops to 60 hours on 24 machines. In terms of parameter size, our result is two orders of magnitude larger than recently-set LDA records in the literature, which involved models with 10s of billions of parameters and typically used massive industrial-scale clusters [11, 21, 1, 12]; our data size is also at least comparable or 1 order of magnitude larger than those same works[1]. As for throughput, our system is able to sample 50 million documents (of average length 200 tokens) worth of latent topic indicators per hour per 20-core machine; which compares favorably to previously reported results: PLDA+ with roughly 1200 documents per hour per machine using the standard collapsed Gibbs sampler, and YahooLDA with roughly 2 million documents per hour per 8-core machine.

Overall, LightLDA benefits both from a highly efficient MCMC sampler built on a new proposal scheme, and a highly efficient distributed architecture and implementation built on Petuum. It represents a truly lightweight realization (hence its name LightLDA) of a massive ML program, which we hope will be easily accessible to ordinary users and researchers with modest resources. Compared to using alternative platforms like Spark and Graphlab that also offer highly sophisticated data- or model- parallel systems, or designing bespoke ground-up solutions like PLDA and YahooLDA, we suggest that our intermediate approach that leverages both simple-but-critical algorithmic innovation and lightweight ML-friendly system platforms stands as a highly cost-effective solution to Big ML.

## 2 Challenges and Related Work

Latent Dirichlet Allocation (LDA) [4] is used to recover semantically coherent topics from document corpora, as well as to discover the topic proportions within each document. LDA is a probabilistic generative model of documents, where each document is represented as a mixture over latent topics, and where each topic is characterized by a distribution over words. To find the most plausible topics and document-topic assignments, one must infer the posterior distribution of the LDA parameters (*word-topic* distributions and *doc-topic* distributions), by using either a variational- or sampling-based inference algorithm. We focus on sampling-based algorithms, because they yield very sparse updates that make them well-suited to settings with a massive number of topics $K$.

Much research has been invested into scaling Latent Dirichlet Allocation (LDA) to ever-larger data and model sizes; existing papers usually show an algorithmic focus (i.e. better LDA inference algorithm speed) or a systems focus (i.e. better software to execute LDA inference on a distributed cluster) — or even both foci at once. Recent large-scale implementations of LDA [12, 1, 21, 11] demonstrate that training is feasible on big document corpora (up to billions of docs) using large, industrial-scale clusters with thousands to tens of thousands of CPU cores. One reason why these implementations require large clusters is that they employ either the SparseLDA inference algorithm [22] or the (slower) original collapsed Gibbs sampler inference algorithm [6]; in either case the inference algorithm is a limiting factor to efficiency. We directly address this bottleneck by developing a new $\mathcal{O}(1)$-per-token Metropolis-Hastings sampler that is nearly an order of magnitude faster than SparseLDA — which allows us to process big corpora on a smaller cluster in a reasonable amount of time. We note that the recently developed AliasLDA algorithm [10] provides an alternative solution to the SparseLDA bottleneck. However, on the one hand, AliasLDA's computational complexity is $\mathcal{O}(K_d)$, so it is not good at processing longer documents such as web pages (particularly because the doc-topic tables are dense in the initial iterations, so $K_d$ is large); on the other hand, the AliasLDA paper only describes a single-machine implementation, so it is unclear if it scales well to the distributed setting (especially considering AliasLDA's high space complexity, $\mathcal{O}(K)$ for each word's alias table). In this paper, we demonstrate that our Metropolis-Hastings sampler in all aspects converges faster than AliasLDA in the single-machine setting, which led us to not use AliasLDA.

---
[1][21] used 4.5 billion tokens, while [11] used 5 billion short documents of unspecified length.

The abovementioned large-scale LDA papers differ substantially in the degree to which they employ data-parallelism (splitting documents over machines) versus model-parallelism (splitting the *word-topic* distributions over machines). While a full treatment of each paper's contribution is not possible within the confines of this related work section, at a high level, YahooLDA [1] and parameter-server-based implementations [11] treat the *word-topic* distributions as globally-shared, in that the inference algorithm is agnostic to how the *word-topic* distributions are physically laid out across machines. More importantly, they schedule inference computations on token topic indicators $z_{di}$ in a document-centric manner, and we would therefore classify them as data-parallel-only implementations. As a consequence, we do not expect either [1] or [11] to handle very large topic models with over 1 trillion parameters (the largest reported result was 10 billion parameters in [11]). In particular, [11] assumes that once the entire corpus has been distributed to sufficiently many machines, the local documents on each machine will only activate a small portion of the LDA model, and therefore the memory required by each machine will not be too large. Consequently, their design cannot handle large topic models without a large compute cluster.

On the other hand, PLDA+ [12] and Peacock [21] additionally group token topic indicators $z_{di}$ according to their word $w_{di}$; this is beneficial because it reduces the proportion of the *word-topic* distributions that must be held at each worker machine — effectively, model-parallelism on top of data-parallelism. In particular, [21] adopted a grid-like model-parallel partitioning strategy, that requires communicating both the training data and LDA model to worker machines (though this requires additional network overhead compared to our design). Also notable is the pipelined design in [12], which only requires workers to hold a small portion of the model in memory; however, their system used an outdated, slow Gibbs sampler, while aspects of their data placement and scheduler are unsuitable for extremely large data and models. Specifically, their word-bundling strategy relies on an inverted index representation of training data, which doubles the memory used by documents (and which we cannot afford since memory is always at a premium in large-scale LDA). Our own LightLDA adopts a different data-and-model-parallel strategy to maximize memory and CPU efficiency: we slice the *word-topic* distributions (the LDA model) in a structure-aware model-parallel manner [9, 24], and we fix blocks of documents to workers while transferring needed model parameters to them via a bounded-asynchronous data-parallel scheme [8]. This allowed us to train a 1-trillion-parameter LDA model over 1 billion documents using as few as 8 machines, while still enjoying proportional speedups when additional machines are available.

## 3 Structure-Aware Model Parallelism for LDA

When training LDA on web-scale corpora with billions of documents, using 100s of thousands of topics can significantly improve the learned model — one reason being that very large corpora may contain numerous small, niche topics (the "long tail"), which would go undetected when the model only contains a few thousand topics [21]. However, for a topic model with up to a million topics, the resulting *word-topic* distributions (the model) may contain trillions of parameters, because web-scale corpora can easily contain millions of unique words. Even though the LDA model is sparse in practice (i.e. many of the parameters are zero), a trillion-parameter model is nevertheless nearly two orders of magnitude larger than recently reported results [12, 1, 21, 11]; in fact, there is evidence that some existing distributed implementations may not scale to such large models, due to issues with the way the model is partitioned over workers — for example, the system design may naively assume that a worker's documents will never touch more than a small fraction of the model, but this has been shown to be untrue on some corpora [9]. Thus, in addition to partitioning documents over worker machines (which all recent distributed LDA implementations already perform in some way [12, 1, 21, 11]), we must simultaneously partition the model in a conservative fashion, in order to ensure that the workers never run out of memory. We refer to this process as structure-aware model parallelism [9, 24].

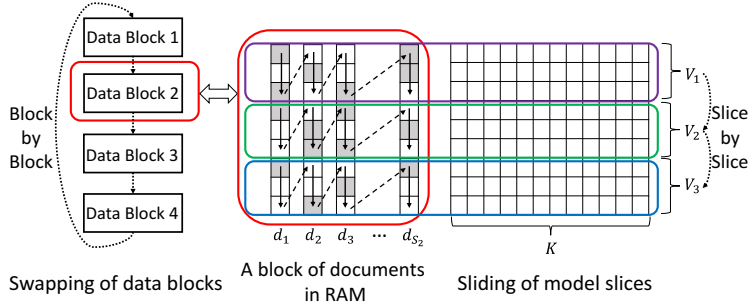

Figure 1: Structure-aware model parallelism in LDA. The arrows across documents $d_1, d_2, \ldots$ indicate the token sampling order — observe that we sample tokens $z$ associated with words $v$ in the *word-topic table* (model) slice $V_1$, before moving on to tokens corresponding to $V_2$, and so forth. Once all document tokens in the data block have been sampled, the system loads the next data block from disk. Each document is sparse (with respect to the vocabulary): shaded cells indicate that the document has one or more tokens $z$ corresponding to word $v$, whereas white indicate that the document has no tokens corresponding to word $v$ (and are hence skipped over).

Before discussing our model parallel strategy, we briefly review the LDA model to establish nomenclature. LDA assumes the following generative process for each document in a corpus:

- $\varphi_k \sim \text{Dirichlet}(\beta)$: Draw word distributions $\varphi_k$ for each topic $k$.
- $\theta_d \sim \text{Dirichlet}(\alpha)$: Draw topic distributions $\theta_d$ for each document $d$.
- $n_d \sim \text{Poisson}(\gamma)$: For each document $d$, draw its length $n_d$ (i.e., the number of tokens it contains).
- For each token $i \in \{1, 2, \ldots, n_d\}$ in document $d$:
  - $z_{di} \sim \text{Multinomial}(\theta_{di})$: Draw the token's topic.
  - $w_{di} \sim \text{Multinomial}(\varphi_{z_{di}})$: Draw the token's word.

The standard collapsed Gibbs sampler for LDA [6] works as follows: all variables except the token's topic indicator $z_{di}$ are analytically integrated out, and we only need to Gibbs sample $z_{di}$ according to

$$p(z_{di} = k | \text{rest}) \propto \frac{(n_{kd}^{-di} + \alpha_k)(n_{kw}^{-di} + \beta_w)}{n_k^{-di} + \bar{\beta}}, \qquad (1)$$

where $w$ is short for $w_{di}$, $\bar{\beta} := \sum_w \beta_w$, $n_{kd}^{-di}$ is the number of tokens in document $d$ that are assigned to topic $k$ (excluding token $z_{di}$), $n_{kw}^{-di}$ is the number of tokens with word $w$ (across all docs) that are assigned to topic $k$ (excluding $z_{di}$), and $n_k^{-di}$ is the number of tokens (across all docs) assigned to topic $k$ (excluding $z_{di}$). To avoid costly recalculation, these counts (also called "sufficient statistics") are cached as tables, and updated whenever a token topic indicator $z_{di}$ changes. In particular, the set of all counts $n_{kd}$ is colloquially referred to as the *document-topic table* (and serves as the sufficient statistics for $\theta_d$), while the set of all counts $n_{kw}$ is known as the *word-topic table* (and forms the sufficient statistics for $\varphi_k$).

At minimum, any distributed LDA implementation must partition (1) the token topic indicators $z_{di}$ and doc-topic table $n_{kd}$ (collectively referred to as the data), as well as (2) the *word-topic table* $n_{kw}$ (the model). When an LDA sampler is sampling a token topic indicator $z_{di}$, it needs to see row $n_{kw_{di}}$ in the *word-topic table* (as well as all of document $d$). However, naive partitioning can lead to situations where some machines touch a large fraction of the *word-topic table*: suppose we sampled every document's tokens in sequence, then the worker would need to see all rows in *word-topic table* corresponding to words in the document. Using our fast Metropolis-Hastings sampler (described in the following section), each worker machine can sample thousands of documents per second (assuming hundreds of tokens per document); furthermore, we have empirically observed that a few million documents (out of billions in our web-scale corpus) is sufficient to activate almost the entire *word-topic table*. Thus, the naive sequence just described would rapidly swap parts of the *word-topic table* (which could be terabytes in size) in and out of each worker, generating a prohibitive amount of network communication.

Our structure-aware model parallel approach is meant to resolve the conflict between fast LDA sampling and limited local memory at each worker; it is similar in spirit to block coordinate descent algorithms. Data blocks are generated prior to running the LDA sampler, and we note that it is cheap to determine which vocabulary words are instantiated by each block. This information is attached as meta-data to the block. As shown in Figure 1, when we load a data block (and its meta-data) into local memory (denoted by the red rectangle), we choose a small set of words (say $V_1$ in the figure) from the block's local words. The set of words is small enough that the corresponding rows $n_{\cdot,w_{di}}$ in *word-topic table* can be held in the worker machine's local memory – we call this set of rows a "model slice". Our system fetches the model slice over the network, and the sampler only samples those tokens in the block that are covered by the fetched slice; all other tokens are not touched. In this manner, the system only maintains a thin model slice in local memory, and re-uses it for all documents in the current data block. Once all tokens covered by the slice have been sampled, the system fetches the next model slice over the network (say $V_2$), and proceeds to sample the tokens covered by it. In this manner (similar to sliding windows in image processing or the TCP/IP protocol), the system processes all tokens in a data block, one slice at a time, before finally loading the next block from disk. This swapping of blocks to and from disk is essentially out-of-core execution.

In addition to keeping worker memory requirements low, this structure-aware model parallelism also mitigates network communication bottlenecks, in the following ways: (1) workers do not move onto the next model slice until all tokens associated with the current slice have been sampled, hence we do not need to apply caching and eviction strategies to the model (which could incur additional communication, as model slices are repeatedly swapped in and out); (2) since the data-model slices are static and unchanging, we pipeline their loading (data blocks from disk, model slices from a central parameter server) to hide network communications latency.

On a final note, we point out that our structure-aware model parallel strategy "sends the model to the data", rather than the converse. This is motivated by two factors: (1) the data (including tokens $w_{di}$ and corresponding topic indicators $z_{di}$) is much larger than the model (even when the model has 1 trillion parameters); (2) as the sampler converges, the model gets increasingly sparse (thus lowering communication), while the data size remains constant. We observe that other distributed LDA designs have adopted a "send data to model" strategy [21], which is costly in our opinion.

## 4 Fast Sampling Algorithm for LDA

As just discussed, the purpose of structure-aware model-parallelism is to enable very large, trillion-parameter LDA models to be learned from billions of documents even on small clusters; furthermore, bounded-asynchronous data-parallel schemes using parameter servers can also reduce the cost of network synchronization and communication (as noted in other distributed LDA papers; see [12, 1, 21, 11]). However, these contributions alone do not allow huge LDA models to be trained *quickly*, and this motivates our biggest contribution: a novel sampling algorithm for LDA, which converges significantly faster than recent algorithms such as SparseLDA [22] and AliasLDA [10]. In order to explain our algorithm, we first review the mechanics of SparseLDA and AliasLDA.

**SparseLDA** SparseLDA [22] exploits the observation that (1) most documents exhibit a small number of topics, and (2) most words only participate in a few topics. This manifests as *sparsity* in both the *doc-topic* and *word-topic* tables, which SparseLDA exploits by decomposing the collapsed Gibbs sampler conditional probability (Eq. 1) into

three terms:

$$p(z_{di} = k|\text{rest}) \propto \underbrace{\frac{\alpha_k \beta_w}{n_k^{-di} + \bar{\beta}}}_{r} + \underbrace{\frac{n_{kd}^{-di} \beta_w}{n_k^{-di} + \bar{\beta}}}_{s} + \underbrace{\frac{n_{kw}^{-di}(n_{kd}^{-di} + \alpha_k)}{n_k^{-di} + \bar{\beta}}}_{t}. \quad (2)$$

When the Gibbs sampler is close to convergence, both the second term $s$ and the third term $t$ will become very sparse (because documents and words settle into a few topics). SparseLDA first samples one of the three terms $r$, $s$ or $t$, according to their probability masses summed over all $k$ outcomes Then, SparseLDA samples the topic $k$ conditioned upon which term $r$, $s$ or $t$ was chosen. If $s$ or $t$ was chosen, then sampling the topic $k$ takes $\mathcal{O}(K_d)$ or $\mathcal{O}(K_w)$ time respectively, where $K_d$ is the number of topics document $d$ contains, and $K_w$ is the number of topics word $w$ belongs to. The amortized sampling complexity of SparseLDA is $\mathcal{O}(K_d + K_w)$, as opposed to $\mathcal{O}(K)$ for the standard collapsed Gibbs sampler.

**AliasLDA** AliasLDA [10] proposes an alternative decomposition to the Gibbs sampling probability:

$$p(z_{di} = k|\text{rest}) \propto \underbrace{\frac{n_{kd}^{-di}(n_{kw}^{-di} + \beta_w)}{n_k^{-di} + \bar{\beta}}}_{u} + \underbrace{\frac{\alpha_k(n_{kw} + \beta_w)}{n_k + \bar{\beta}}}_{v}. \quad (3)$$

AliasLDA pre-computes an alias table [20] for the second term, which allows it to be sampled in $\mathcal{O}(1)$ time via Metropolis-Hastings. By re-using the table over many tokens, the $\mathcal{O}(K)$ cost of building the table is also amortized to $\mathcal{O}(1)$ per token. The first term $u$ is sparse (linear in $K_d$, the current number of topics in document $d$), and can be computed in $\mathcal{O}(K_d)$ time.

## 4.1 Metropolis-Hastings sampling

We have just seen that SparseLDA and AliasLDA achieve $\mathcal{O}(K_d + K_w)$ and $\mathcal{O}(K_d)$ amortized sampling time per token, respectively. Such accelerated sampling is important, because we simply cannot afford to sample token topic indicators $z_{di}$ naively; the original collapsed Gibbs sampler (Eq. 1) requires $\mathcal{O}(K)$ computation per token, which is clearly intractable at $K = 1$ million topics. SparseLDA reduces the sampling complexity by exploiting the sparsity of problem, while AliasLDA harnesses the alias approach together with the Metropolis-Hastings algorithm [14, 7, 19, 3]. Our LightLDA sampler also turns to Metropolis-Hastings, but with new insights into proposal distribution design, which is most crucial for high performance. We show that the sampling process can be accelerated even further with a well-designed proposal distribution $q(\cdot)$ to the true LDA posterior $p(\cdot)$.

A well-designed proposal $q(\cdot)$ should speed up the sampling process in two ways: (1) drawing samples from $q(\cdot)$ will be much cheaper than drawing samples from $p(\cdot)$; (2) the Markov chain should mix quickly (i.e. requires only a few steps). What are the trade-offs involved in constructing a good proposal distribution $q(\cdot)$ for $p(\cdot)$? If $q(\cdot)$ is very similar to $p(\cdot)$, then the constructed Markov chain will mix quickly — however, the cost of sampling from $q(\cdot)$ might end up as expensive as sampling from $p(\cdot)$ itself. On the contrary, if $q(\cdot)$ is very different from $p(\cdot)$, we might be able to sample from it cheaply — but the constructed Markov chain may mix too slowly, and require many steps for convergence. To understand this trade-off, consider the following extremes:

**Uniform Distribution Proposal** Suppose we choose $q(\cdot)$ to be the uniform distribution. The MH algorithm will propose the next state $t \sim Unif(1, ..., K)$, and accept the proposed state with probability $\min\{1, \frac{p(t)}{p(s)}\}$. Obviously, sampling from a uniform distribution is very cheap and can be done in $\mathcal{O}(1)$ time; however, the uniform distribution is non-sparse and therefore extremely far from $p(\cdot)$, thus it needs many MH steps to mix.

**Full Conditional Distribution Proposal** We could instead choose $p(\cdot)$ itself as the proposal distribution $q(\cdot)$. The MH algorithm proposes the next step $t$ with probability $p(t)$, and accepts it with probability $\min\{1, \frac{p(t)p(s)}{p(s)p(t)}\} = 1$; i.e. the algorithm accepts all proposals. Sampling from $q(\cdot)$ obviously costs as much as $p(\cdot)$, but mixing is very fast because all proposals are accepted.

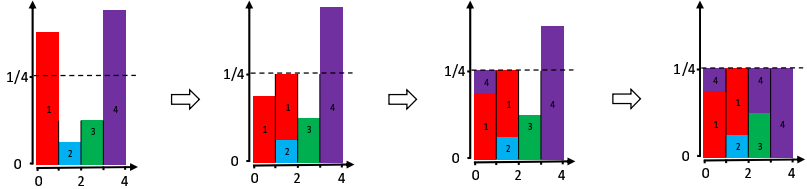

Figure 2: An example showing how to build an alias table. This procedure transforms a non-uniform sampling problem into a uniform sampling one. The alias table maintains the mass of each bin and can be re-used once constructed. More details about the alias method can be found in [13].

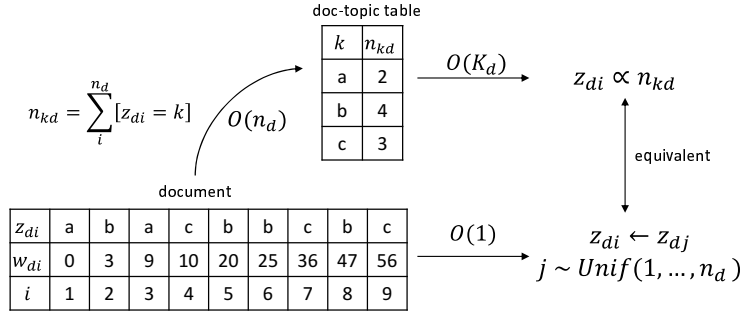

Figure 3: Illustration of how we sample the doc-proposal in $\mathcal{O}(1)$ time, without having to construct an alias table.

## 4.2 Cheap Proposals by Factorization

To design an MH algorithm that is cheap to draw from, yet has high mixing rate, we adopt a *factorized* strategy: instead of a single proposal, we shall construct a set of $\mathcal{O}(1)$ proposals, and alternate between them. To construct these proposals, let us begin from the true conditional probability of token topic indicator $z_{di}$:

$$p(k) = p(z_{di} = k|\text{rest}) \propto \frac{(n_{kd}^{-di} + \alpha_k)(n_{kw}^{-di} + \beta_w)}{n_k^{-di} + \bar{\beta}}. \tag{4}$$

Observe that it can be decomposed into two terms:

$$q(z_{di} = k|\text{rest}) \propto \underbrace{(n_{kd} + \alpha_k)}_{\text{doc–proposal}} \times \underbrace{\frac{n_{kw} + \beta_w}{n_k + \bar{\beta}}}_{\text{word–proposal}}. \tag{5}$$

Even if we exploit sparsity in both terms, sampling from this conditional probability costs at least $\mathcal{O}(\min(K_d, K_w))$ — can we do better? We observe that the first term is document-dependent but word-independent, while the second term is doc-independent but word-dependent. Furthermore, it is intuitive to see that the most probable topics are those with high probability mass from both the doc-dependent term and the word-dependent term; hence, either term alone can serve as a good proposal $q$ — because if $p$ has high probability mass on topic $k$, then either term will also have high probability mass on $k$ (though the converse is not true). Just as importantly, the alias method [13] (also used in AliasLDA [10]) can be applied to both terms, reducing the cost of sampling from either proposal to just $\mathcal{O}(1)$ amortized time per token (where the cost of constructing the alias table is getting amortized). We now discuss the proposals individually.

**Word-Proposal for Metropolis-Hastings** Define $p_w$ as the word-proposal distribution

$$p_w(k) \propto \frac{n_{kw} + \beta_w}{n_k + \bar{\beta}}. \tag{6}$$

The acceptance probability of state transition $s \to t$ is

$$\min\{1, \frac{p(t)p_w(s)}{p(s)p_w(t)}\}. \quad (7)$$

Let $\pi_w := \frac{p(t)p_w(s)}{p(s)p_w(t)}$, we can show that

$$\pi_w = \frac{(n_{td}^{-di} + \alpha_t)(n_{tw}^{-di} + \beta_w)(n_s^{-di} + \bar{\beta})(n_{sw} + \beta_w)(n_t + \bar{\beta})}{(n_{sd}^{-di} + \alpha_s)(n_{sw}^{-di} + \beta_w)(n_t^{-di} + \bar{\beta})(n_{tw} + \beta_w)(n_s + \bar{\beta})}. \quad (8)$$

Once $t \sim p_w(t)$ is sampled, the acceptance probability can be computed in $\mathcal{O}(1)$ time, as long as we keep track of all sufficient statistics $n.$ during sampling. Intuitively, $\pi_w$ is high (relative to topic $s$) whenever the proposed topic $t$ is either (1) popular within document $d$, or (2) popular for the word $w$. Since the word-proposal tends to propose topics $t$ which are popular for word $w$, using the word-proposal will have explore the state space of $p(k)$. To sample from $p_w$ in $\mathcal{O}(1)$, we use alias table similar to [10]. As illustrated by Figure 2, the basic idea of the alias approach is to transform a non-uniform distribution into a uniform distribution (i.e., alias table). Since the alias table will be re-used in MH sampling, the transformation cost gets amortized to $\mathcal{O}(1)$ [2].

Although the alias approach has low $\mathcal{O}(1)$ amortized time complexity, its space complexity is still very high, because the alias table for each word's proposal distribution stores $2K$ values: the splitting point of each bin and the alias value above that splitting point; this becomes prohibitive if we need to store a lot of words' alias tables. Our insight here is that the alias table can be sparsified; specifically, we begin by decomposing $p_w = \frac{n_{kw}}{n_k+\beta} + \frac{\beta_w}{n_k+\beta}$. We then draw one of the two terms, with probability given by their masses (this is known as a mixture approach[3]). If we draw the first term, we use a pre-constructed alias table (created from $n_{kw}$, specific to word $w$) to pick a topic, which is sparse. If we draw the second term, we also use a pre-constructed alias table (created from $n_k$, common to all words $w$ and thus amortized over all $V$ words) to pick a topic, which is dense. In this way, we reduce both the time and space complexity of building word $w$'s alias table to $\mathcal{O}(K_w)$ (the number of topics word $w$ participates in).

**Doc-Proposal for Metropolis Hastings** Define $p_d$ as the doc-proposal distribution

$$p_d(k) \propto n_{kd} + \alpha_k. \quad (9)$$

The acceptance probability of state transition $s \to t$ is

$$\min\{1, \frac{p(t)p_d(s)}{p(s)p_d(t)}\}. \quad (10)$$

Let $\pi_d := \frac{p(t)p_d(s)}{p(s)p_d(t)}$, we can show that

$$\pi_d = \frac{(n_{td}^{-di} + \alpha_t)(n_{tw}^{-di} + \beta_w)(n_s^{-di} + \bar{\beta})(n_{sd} + \alpha_s)}{(n_{sd}^{-di} + \alpha_s)(n_{sw}^{-di} + \beta_w)(n_t^{-di} + \bar{\beta})(n_{td} + \alpha_t)}. \quad (11)$$

As with the word-proposal, we see that the doc-proposal accepts whenever topic $t$ is popular (relative to topic $s$) within document $d$, or popular for word $w$. We decompose $p_d(k) \propto \frac{n_{kd}}{n_d+\alpha} + \frac{\alpha_k}{n_d+\alpha}$ just like the word-proposal, except that when we pick the first term, we do not even need to explicitly build the alias table — this is because the document token topic indicators $z_{di}$ serve as an alias table. Specifically, the first term $n_{kd}$ counts the number of times topic $k$ occurs in document $d$, in other words

$$n_{kd} = \sum_{i=1}^{n_d} [z_{di} = k], \quad (12)$$

---

[2] This strategy keeps the Metropolis-Hastings proposal $p_w$ fixed over multiple documents, rather than changing it after every token. This is well-justified, since Metropolis-Hastings allows any proposal (up to some conditions) provided the acceptance probability can be correctly computed. We have already argued in Section 4.2 that the acceptance probability can be computed in $\mathcal{O}(1)$ time by simply keeping track of a few sufficient statistics $n_{..}$.

[3] This mixture strategy was also used by SparseLDA [22], but on the true Gibbs sampling probability rather than an MH proposal.

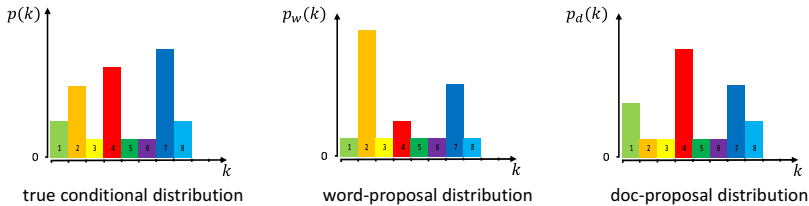

Figure 4: An example that explains why cycling two proposals helps mixing rate. The 2nd bin is a mode of $p(k)$, and $p_d(k)$ is obviously not a good proposal for this mode — but $p_w(k)$ is good at exploring it.

where $[\cdot]$ is the indicator function. This implies that the array $z_{di}$ is an alias table for the unnormalized probability distribution $n_{kd}$, and therefore we can sample from $n_{kd}$ by simply drawing an integer $j$ uniformly from $\{1, 2, \ldots, n_d\}$, and setting $z_{di} = z_{dj}$. Figure 3 uses a toy example to illustrate this procedure. Hence, we conclude that the doc-proposal can be sampled in $\mathcal{O}(1)$ non-amortized time (because we do not need to construct an alias table)[4].

### 4.3 Combining Proposals to Improve Mixing

While either the doc- or word-proposal alone can be used as an as efficient MH algorithm for LDA, in practice many MH-steps (repeatedly sampling each token) are required to produce proper mixing. With a small number of MH-steps, using the word-proposal alone encourages sparsity in word-topic distribution (i.e. each word belongs to few topics) but causes low sparsity in document-topic distributions (i.e. each document contains many topics). Conversely, using the doc-proposal alone with few MH-steps leads to sparsity in document-topic distribution but non-sparse word-topic distributions. Therefore, while either proposal can sample tokens very quickly, they need many MH-steps to mix well.

The key to fast Metropolis-Hastings mixing is a proposal distribution that can quickly explore the state space, and reach all states with high probability (the modes). The word-proposal $p_w(k)$ is good at proposing only its own modes (resulting in concentration of words in a few topics), and likewise for the doc-proposal $p_d(k)$ (resulting in concentration of docs onto a few topics). As Figure 4 shows, with the word-proposal or doc-proposal alone, some modes will never be explored quickly.

How can we achieve a better mixing rate while still maintaining high sampling efficiency? If we look at $p(k) \propto p_w(k) \times p_d(k)$, we see that for $p(k)$ to be high (i.e. a mode), we need either $p_w(k)$ or $p_d(k)$ to be sufficiently large — but not necessarily both at the same time. Hence, our solution is to combine the doc-proposal and word-proposal into a "cycle proposal"

$$p_c(k) \propto p_d(k) p_w(k), \qquad (13)$$

where we construct an MH sequence for each token by alternating between doc- and word-proposal. The results of [19] show that such cycle MH proposals are theoretically guaranteed to converge. By combining the two proposals in this manner, all modes in $p(k)$ will be proposed, with sufficiently high probability, by at least one of the proposals. Another potential benefit of cycling different proposals is that it helps to reduce the auto-correlation among sampled states, thus exploring the state space more quickly.

## 5 Hybrid Data Structures for Power-Law Words

Even with careful data-model partitioning, memory size remains a critical obstacle when scaling LDA to very large numbers of topics. The LDA model, or *word-topic table* $n_{kw}$, is a $V \times K$ matrix, and a naive dense representation would require prohibitive amounts of memory — for example, for $V = K = 1$ million used in this paper's experiments, the model would be 4 terabytes in size assuming 32-bit integer entries. Even with reasonably well-equipped machines with 128 gigabytes of RAM, just storing the matrix in memory would require 32 machines — and in practice, the actual usage is often much higher due to other system overheads (e.g. cache, alias tables, buffers, parameter server).

---
[4]Unlike the word-proposal, $p_d$ changes after every token to reflect the current state of $z_{di}$. Again, this is fine under Metropolis-Hastings theory.

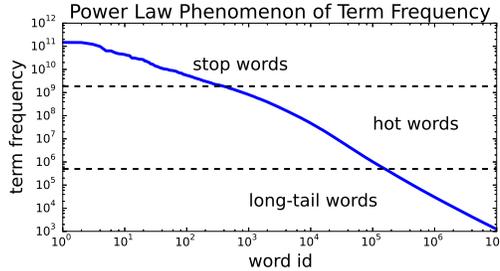

Figure 5: Word frequencies in topic modeling follow a power-law phenomenon (log-log plot). The great difference in the frequency of hot words versus long-tailed words makes selecting the right data structure difficult, as discussed in the text. This plot is obtained from 15 billion web pages, with over than 3000 billion tokens.

A common solution is to turn to sparse data structures such as hash maps. The rationale behind sparse storage is that document words follow a power law distribution (Figure 5). There are two implications: (1) after removing stop words, the term frequency of almost all meaningful words will not exceed the upper range of a 32-bit integer (2,147,483,647); this was measured on a web-scale corpus with 15 billion webpages and over 3000 billion tokens, and only 300 words' term frequencies exceed the 32-bit limit. For this reason, we choose to use 32-bit integers rather than 64-bit ones. (2) Even with several billion documents, it turns out the majority of words occur fewer than $K$ times (where $K$ is the number of topics, up to 1 million in our experiments). This necessarily means that most rows $n_{k,\cdot}$ in the *word-topic table* are extremely sparse, so a sparse row representation (hash maps) will significantly reduce the memory footprint.

However, compared to dense arrays, sparse data structures exhibit poor random access performance, which hurts MCMC algorithms like SparseLDA, AliasLDA and our Metropolis-Hastings algorithm because they all rely heavily on random memory references. In our experiments, using pure hash maps resulted in a several-fold performance loss compared to dense arrays. How can we enjoy low memory usage whilst maintaining high sampling throughput? Our solution is a hybrid data structure, in which *word-topic table* rows corresponding to frequent hot words are stored as dense arrays, while uncommon, long-tail words are stored as open-addressing/quadratic-probing hash tables. In our web-scale corpus with several billion documents, we found that $10\%$ of the vocabulary words are "hot" and cover almost $95\%$ of all tokens in the corpus, while the remaining $90\%$ of vocabulary words are long-tail words that cover only $5\%$ of the tokens. This implies that (1) most accesses to our hybrid *word-topic table* go to dense arrays, which keeps throughput high; (2) most rows of the *word-topic table* are still sparse hash tables[5], which keeps memory usage reasonably low. In our $V = K = 1$ million experiments, our hybrid *word-topic table* used 0.7TB, down from 4TB if we had used purely dense arrays. When this table is distributed across 24 machines, only 30GB per machine is required, freeing up valuable memory for other system components.

## 6  System Implementation

Distributed implementations are clearly desirable for web-scale data: they reduce training time to realistic levels, and most practitioners have access to at least a small distributed cluster. However, existing distributed LDA implementations have only been shown to work at much smaller problem scales (particularly model size), or suggest the use of extremely large compute clusters (sometimes numbering in the thousands of machines) to finish training in acceptable time. What are the challenges involved in solving big LDA problems on just tens of machines? If we want to train on a corpus with billions of documents (each with at least hundreds of tokens) occupying terabytes of space, then on the data-parallel side, simply copying the data from disk to memory will take tens of hours, while transferring the data over the network also takes a similar amount of time. On the model-parallel side, storing 1 trillion parameters (1 million words by 1 million topics) can take up to terabytes of memory — necessitating distributed storage, which in turn requires inter-machine parameter synchronization, and thus high network communication cost. In light of these considerations, our goal is to design an architecture for LightLDA that reduces these data transmission and parameter communication costs as much as possible, thus making execution on small clusters realistic.

---

[5] In order to further improve the throughput of the long tail words, we set the capacity of each hash table to at least two times the term frequency of a long-tail word. This guarantees a load factor that is $\leq 0.5$, thus keeping random access performance high.

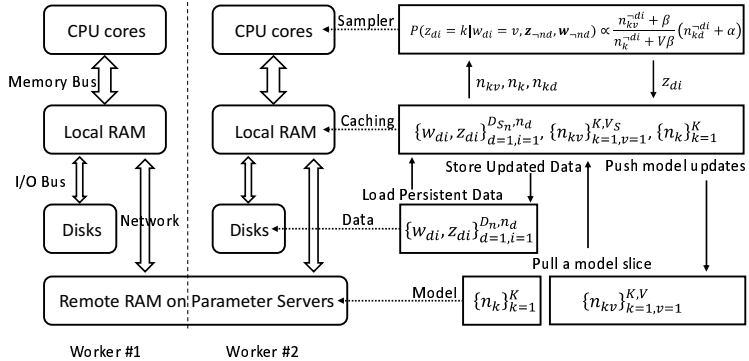

Figure 6: System architecture, data/model placement and logical flow.

**System Overview** We build LightLDA on top of an open-source distributed machine learning system Petuum (www.petuum.org), which provides a general framework for structure-aware model parallelism and bounded-asynchronous data-parallelism in large-scale machine learning. In terms of code, we specifically make use of its parameter server [18, 8, 11] for bounded-asynchronous data-parallelism. A parameter server hides the distributed networking details (such as communication and concurrency control) from users, and provides elegant APIs for the development of distributed machine learning programs — the idea being to let Machine Learning experts focus on algorithmic logic rather than the deep system details. We first introduce the general parameter server idea, and then describe our substantial enhancements to make big LDA models possible on small clusters.

**Parameter Server and Data Placement** At the basic level, a parameter server (PS) presents a distributed shared memory interface [16], where programmers can access any parameter from any machine, agnostic to the physical location of the parameter. Essentially, the PS extends the memory hierarchy of a single machine (Figure 6); storage media closer to the CPU cores has lower access latency and higher transmission bandwidth, but has much smaller capacity. In the PS architecture, each machine's RAM is split into two parts: local RAM for client usage and remote RAM for centralized parameter storage (referred to as the "server" part). These hardware limitations, together with the requirements imposed by big topic model data model, strongly influence the manner in which we run our Metropolis-Hastings algorithm.

We use the PS to store two types of LDA model parameters: the *word-topic table* $\{n_{kv}\}_{k=1,v=1}^{K,V}$, which counts the number of tokens with word $v$ assigned to topic $k$, and a length-$K$ "summary row" $\{n_k\}_{k=1}^{K}$ which counts the total number of tokens assigned to topic $k$ (regardless of word). 32-bit integers are used for the *word-topic table* (using a combination of dense arrays and sparse hash maps; see Section 5), and a 64-bit integer array for the summary row. We observe that as the LDA sampler progresses, the *word-topic table* becomes increasingly sparse, leading to lower network communication costs as time passes. Furthermore Petuum PS supports a bounded-asynchronous consistency model [8], which reduces inter-iteration parameter synchronization times through a staleness parameter $s$ — for LightLDA, which is already a heavily-pipelined design, we found the optimal value to be $s = 1$.

Given that the input data are much larger than the model (and remain unchanged throughout LDA inference), it is unwise to exchange data over the network. Instead, we shuffle and shard the corpus across the disks of all worker machines, and each worker machine only ever accesses the data in its local disk. In Figure 6, $\{w_{di}, z_{di}\}_{d=1,i=1}^{D_n, n_d}$ indicates a shard of training data in the $n$-th worker machine, where $D_n$ represents the number of documents in the $n$-th worker, $n_d$ indicates the number of tokens in document $d$. Each worker's local memory holds (1) the active working set of data $\{w_{di}, z_{di}\}_{d=1,i=1}^{D_{S_n}, n_d}$, and (2) the model $\{n_{kv}\}_{k=1,v=1}^{K, V_S}$ required to sample the current set of tokens (using the Metropolis-Hastings sampler). During sampling, we update the token topic indicators $z_{di}$, and the *word-topic table*. The token-topic pairs ($(w_{di}, z_{di})$ are local to the worker machine and incur no network communication, while the *word-topic table* is stored in the PS therefore requires a background thread for efficient communication.

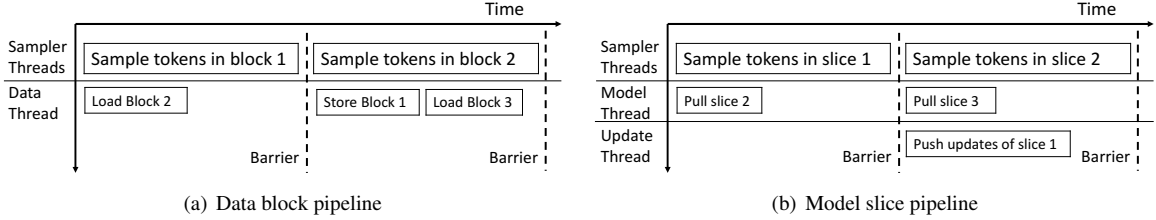

Figure 7: Pipelines to overlap computation, disk I/O and network.

**Token and Topic Indicator Storage** As part of data-parallel execution, each worker machine stores a shard of the corpus on its local disk. For web-scale corpora, each shard may still be very large — hundreds of gigabytes, if not several terabytes — which prohibits loading the entire shard into memory. Thus, we further split each data shard into data blocks, and stream the blocks one at a time into memory (Figure 1 left). In terms of data structures, we deliberately place tokens $w_{di}$ and their topic indicators $z_{di}$ side-by-side, as a vector of $(w_{di}, z_{di})$ pairs rather than two separate vectors for tokens and topic indicators (which was done in [1]). We do this to improve data locality and CPU cache efficiency: whenever we access a token $w_{di}$, we always need to access its topic indicator $z_{di}$, and the vector-of-pairs design directly improves locality. One drawback to this design is extra disk I/O, from reading/writing the (read-only) tokens $w_{di}$ to disk every time a data shard gets swapped out. However, disk I/O can always be masked via pipelined reads/writes, done in the background while the sampler is processing the current shard.

We point out that our streaming and disk-swapping (out-of-core) design naturally facilitates fault tolerance in the following way: if we perform data swapping to disk via atomic file overwrites, then whenever the system fails, it can simply resume training via warm-start: read the swapped-to-disk model, re-initialize the word-topic and doc-topic tables, and carry on. In contrast, for LDA systems like PLDA+ [12] and YahooLDA [1] to have fault recovery, they would require periodic dumps of the data and/or model — but this incurs a nontrivial cost in the big data/model scenarios that we are trying to address.

**Tuning the Structure-Aware Model Parallelization** In Section 3, we introduced the high-level idea of structure-aware model parallelization as applied to LDA, and there are still a number of improvements that can be employed to improve its efficiency. We present the most notable ones:

1. After completing a data block or a model slice, a worker machine's CPU cores need to wait for the next data block/model slice to be loaded from disk/network respectively. We eliminate this I/O latency via pipelining (Figure 7), though we caution that perfect pipelining requires careful parameter configuration (taking into consideration the throughput of samplers, size of data blocks, size of model slices).

2. To prevent data load imbalances across model slices, we generate model slices by sorting the vocabulary by word frequencies, and then shuffling the words. In this manner, each slice will contain both hot words and long tail words, improving load balance.

3. To eliminate unnecessary data traversal, when generating data blocks, we sort token-topic pairs $(w_{di}, z_{di})$ according to $w_{di}$'s position in the shuffled vocabulary, ensuring that all tokens belonging to the same model slice are actually contiguous in the data block (see Figure 1). This sorting only needs to be performed once, and is very fast on data processing platforms like Hadoop (compared to the LDA sampling time). We argue that this is more efficient than the "word bundle" approach in PLDA+ [12], which uses an inverted index to avoid data traversal, but at the cost of doubling data memory requirements.

| DATASET | V | L | D | L/V | L/D |
|---|---|---|---|---|---|
| NYTIMES | 101636 | 99542125 | 299752 | 979 | 332 |
| PUBMED | 141043 | 737869083 | 8200000 | 5231 | 90 |
| BING WEBC | 1000000 | 200B | 1.2B | 200000 | 167 |

Table 1: Experimental datasets and their statistics. V denotes vocabulary size, L denotes the number of training tokens, D denotes the number of documents, L/V indicates the average number of occurrences of a word, L/D indicates the average length of a document. For the Bing web chunk data, 200B denotes 200 billion.

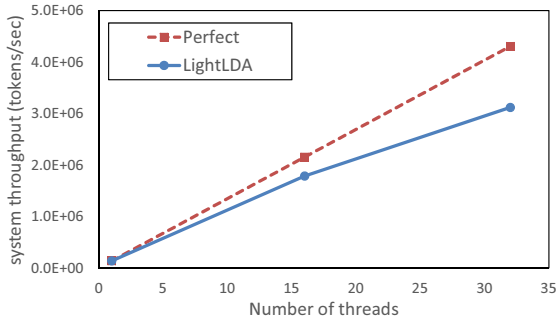

Figure 8: Intra-node scalability of LightLDA, using 1, 16, 32 threads on a single machine (no network communication costs).

4. We use a bounded asynchronous data parallel scheme [8] to remove the network waiting time occurring at the boundary of adjacent iterations. Note that, to pre-fetch the first slice of model for the next data block, we do not need to wait for the completion of the sampling on the current data block with the last slice of the model. We call this Stale Synchronous Parallel (SSP), which is shown to outperform the well known Bulk Synchronous Parallel (BSP) programming model.

**Multi-thread Efficiency** Our sampler is embarrassingly parallel within a single worker machine. This is achieved by splitting the in-memory data block into disjoint partitions (to be sampled by individual threads), and sharing the in-memory model slice amongst the threads. Furthermore, we make the shared model slice immutable, and delay all model updates locally before sending them to be aggregated at the parameter server. By keeping the model slice immutable, we avoid concurrency issues such as race conditions and locking, thus achieving near-linear intra-node scalability. While delaying model updates theoretically slows down the model convergence rate, in practice, it eliminates concurrency issues and thus increases sampler throughput, easily outweighing the slower convergence rate.

Modern server-grade machines contain several CPU sockets (each CPU houses many physical cores) which are connected to separate memory banks. While these banks can be addressed by all CPUs, memory latencies are much longer when accessing remote banks attached to another socket — in other words, Non-Uniform Memory Access (NUMA). In our experiments, we have found NUMA effects to be fairly significant, and we partially address them through tuning sampling parameters such as the number of Metropolis-Hastings steps (which influences CPU cache hit rates, and mitigates NUMA effects). That said, we believe proper NUMA-aware programming is a better long-term solution to this problem. Finally, we note that setting core affinities for each thread and enabling hardware hyper-threading on Intel processors can be beneficial; we observed a $30\%$ performance gain when employing both.

## 7 Experimental Results

We demonstrate that LightLDA is able to train much larger LDA models on similar or larger data sizes than previous LDA implementations, using much fewer machines — due to careful data-model slicing, and especially our new Metropolis-Hastings sampler that is nearly an order of magnitude faster than SparseLDA and AliasLDA. We use several datasets (Table 7), notably a Bing "web chunk" dataset with 1.2 billion webpages (about 200 billion tokens in total). Our experiments show that (1) our distributed implementation of LightLDA has near-linear scalability in the number of cores and machines; (2) the LightLDA Metropolis-Hastings sampler converges significantly faster than the state-of-the-art SparseLDA and AliasLDA samplers (measured in a single-threaded setting); (3) most importantly, LightLDA enables very large data and model sizes to be trained on as few as 8 machines.

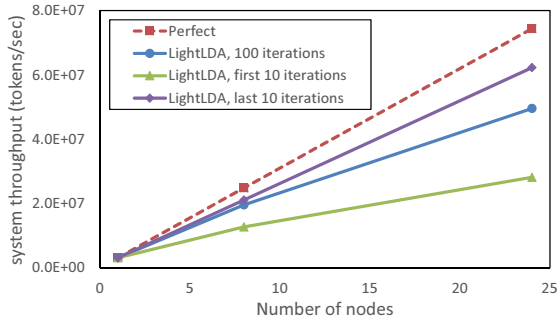

Figure 9: Inter-node scalability of LightLDA, using 1, 8, 24 machines.

## 7.1 Scalability of Distributed LightLDA

In these experiments, we establish that our LightLDA implementation scales almost linearly in computational resources. We begin with intra-node, single-machine multi-threaded scalability: to do this, we restrict the web chunk dataset to the 50,000 most frequent words, so that the model can be held in the RAM of a single machine. We then run LightLDA with 1, 16, 32 threads (on a 32-core machine), training a model with 1 million topics on the web chunk dataset. We record the number of tokens sampled by the algorithm over 2 iterations, and plot it in Figure 8. The figure shows that LightLDA exhibits nearly linear scaling inside a single machine.

Next, we demonstrate that LightLDA scales in the distributed setting, with multiple machines. Using the same web chunk dataset with $V = 50,000$ vocabulary words and $K = 1$ million topics, we run LightLDA on 1, 8, 24 machines (using 20 threads per machine, and 256GB RAM per machine; such configurations can be readily purchased from cloud compute providers), and set the Petuum parameter server to use a staleness value of $1^6$. This time, we record the number of tokens sampled over 100 iterations, and compute the average token throughput over the first 10 iterations, the last 10 iterations, and all 100 iterations. These results are plotted in Figure 9, and we note that scalability is poor in the first 10 iterations, but close to perfect in the last 10 iterations. This is because during the initial iterations, the LDA model is still very dense (words and documents are assigned to many topics, resulting in large model size; see Figure 10(c)), which makes the system incur high network communication costs (Figure Figure 10(b)) — enough to saturate our cluster's 1 Gbps Ethernet. In this situation, our pipelining (explained in Figure 7) does not mask the extra communication time, resulting in poor performance. However, after the first 10 iterations or so, the model becomes sparse enough for pipelining to mask the communication time, leading to near-perfect inter-node scalability. We conjecture that upgrading to 10 Gbps Ethernet (frequently seen in industrial platforms and cloud computing services) or better will eliminate the initial bottleneck during the first few iterations.

Finally, we demonstrate that LightLDA is able to handle very large model sizes: we restore the full $V = 1$ million vocabulary of the web chunk data and $K = 1$ million topics, yielding a total of 1 trillion model parameters on 200 billion tokens. We then run LightLDA on this large dataset and model using 8 and 24 machines, and plot the log-likelihood curves in (Figure 10(a)). Convergence is observed within 2 days on 24 machines (or 5 days on 8 machines), and it is in this sense that we claim big topic models are now possible on modest compute clusters.

One might ask if overfitting happens on such large models, given that the number of parameters (1 trillion) is larger than the number of tokens (200 billion). We point out that (1) there is evidence to show that large LDA models can improve ad prediction tasks [21], and (2) the converged LDA model is sparse, with far fewer than 200 billion nonzero elements. As evidence, we monitored the number of non-zero entries in *word-topic table* during the whole training process, and observed that after 100 iterations, the model had only 2 billion non-zero entries (which is 1% of 200 billion tokens).

---

[6]This staleness setting is part of Stale Synchronous Parallel data-parallelism [8]. A positive staleness value lets LightLDA mask the delay between iterations due to transferring model parameters over the network, yielding higher throughput at almost no cost to quality.

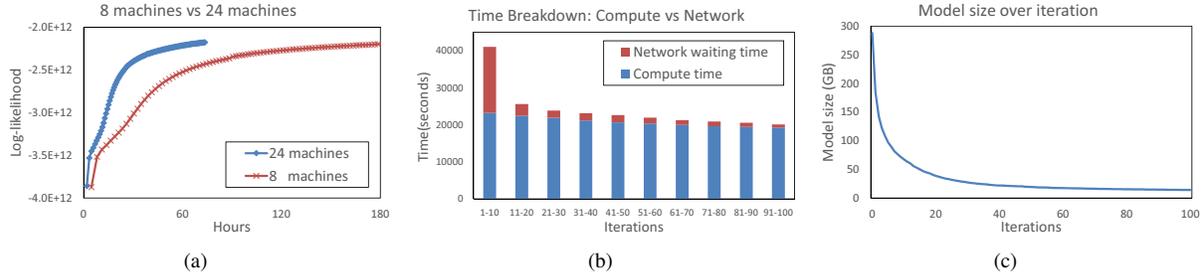

Figure 10: (a) Log-likelihood over running time with 8 and 24 machines, on Bing web chunk dataset, using $V = K = 1$ million (1 trillion parameters). (b) Compute time v.s. Network waiting time, as a function of iteration number. Observe that communication costs are significantly higher during the first 10 iterations, when the model is still dense. (c) Model size (non-zero entries in *word-topic table*) versus iteration number.

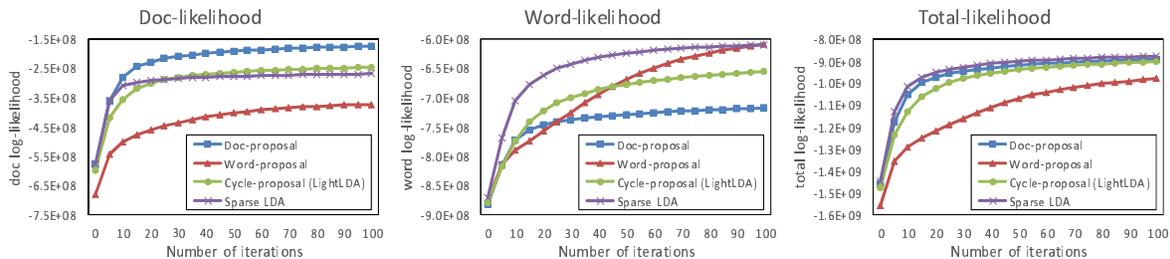

Figure 11: Performance of different LightLDA Metropolis-Hastings proposals, on the NYTimes data set with $K = 1000$ topics.

## 7.2 LightLDA Algorithm versus Baselines

We also want to establish that our Metropolis-Hastings algorithm converges faster than existing samplers (SparseLDA and AliasLDA) to a high quality answer. Using a single computational thread, we ran LightLDA, SparseLDA and AliasLDA on two smaller datasets (NYTimes, PubMed) using $K = 1,000$ or $10,000$ topics, and plotted the log-likelihood versus running time in Figure 12. We note that AliasLDA consistently performs better than SparseLDA (as reported in [10]), and that LightLDA is around 3 to 5 times as fast as AliasLDA. To better understand the performance differences, we also plot the time taken by the first 100 iterations of each algorithm in Figure 13. In general, LightLDA has a consistently low iteration time[7]. AliasLDA is significantly faster than SparseLDA on datasets with short documents (PubMed[8]), but is only marginally faster on longer documents (NYTimes).

Although LightLDA is certainly faster per-iteration than SparseLDA and AliasLDA, to obtain a complete picture of convergence, we must plot the log-likelihood versus each iteration (Figure 14). We observe that SparseLDA makes the best progress per iteration (because it uses the original conditional Gibbs probability), while LightLDA is usually close behind (because it uses simple Metropolis-Hastings proposals). AliasLDA (another Metropolis-Hastings algorithm) is either comparable to LightLDA (on NYTimes) or strictly worse (on Pubmed). Because the time taken for each LightLDA and AliasLDA algorithm is very short (Figure 13), in terms of convergence *per time*, LightLDA and AliasLDA are significantly faster than SparseLDA (with LightLDA being the faster of the two).

---
[7]Unlike the web chunk experiments, there is no inter-machine communication bottleneck, and therefore LightLDA does not experience slow initial iterations.

[8]Note that we use the whole PubMed data set in this experiment, whereas the AliasLDA paper [10] only considered 1% of the total PubMed data.

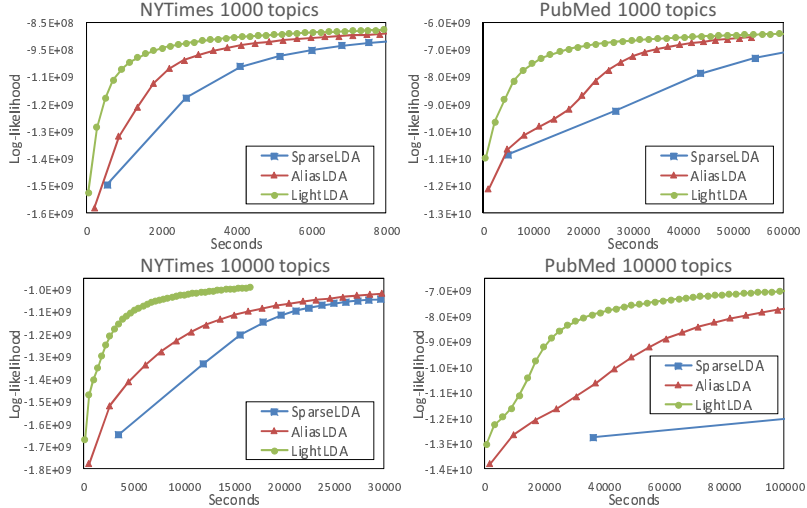

Figure 12: Log-likelihood versus running time for SparseLDA, AliasLDA, LightLDA.

Finally, we argued that the LightLDA "cycle proposal", which alternates between the doc-proposal and word-proposal, would be highly effective at exploring the model space, and thus a better option than either proposal individually. To demonstrate this, in Figure 11, we plot the performance of the full LightLDA cycle proposal versus the doc- and word-proposals alone, as well as SparseLDA (which represents the gold standard for quality, being a non-Metropolis-Hastings Gibbs sampler). The plots are subdivided into (1) full likelihood, (2) document likelihood, and (3) word likelihood ([18] explains this decomposition in more detail). Intuitively, a high doc-likelihood shows that the learned topics compactly represent all documents, while a high word-likelihood indicates that the topics compactly represent the vocabulary. Although the doc- and word-proposals appear to do well on total likelihood, in actual fact, the word-proposal fails to maximize the doc-likelihood, while the doc-proposal fails to maximize the word-likelihood. In contrast, the full LightLDA cycle proposal represents both documents and vocabulary compactly, and is nearly as good in quality as SparseLDA.

## 8  Conclusions

We have implemented a distributed LDA sampler, LightLDA, that enables very large data sizes and models to be processed on a small compute cluster. LightLDA features significantly improved sampling throughput and convergence rate via a (surprisingly) fast $\mathcal{O}(1)$ Metropolis-Hastings algorithm, and allows even small clusters to tackle very large data and model sizes thanks to careful structure-aware model-parallelism and bounded-asynchronous data-parallelism on the Petuum framework. A hybrid data structure is used to simultaneously maintain good performance and memory efficiency, providing a balanced trade-off. On a future note, we believe the Metropolis-Hastings decomposition in our sampler can be successfully applied to inference of other graphical models, alongside the general concepts of structure-aware model-parallelism and bounded-asynchronous data-parallelism. It is our hope that more ML applications can be run with big data and model sizes on small, widely-available clusters, and that this work will inspire current and future development of large-scale ML systems.

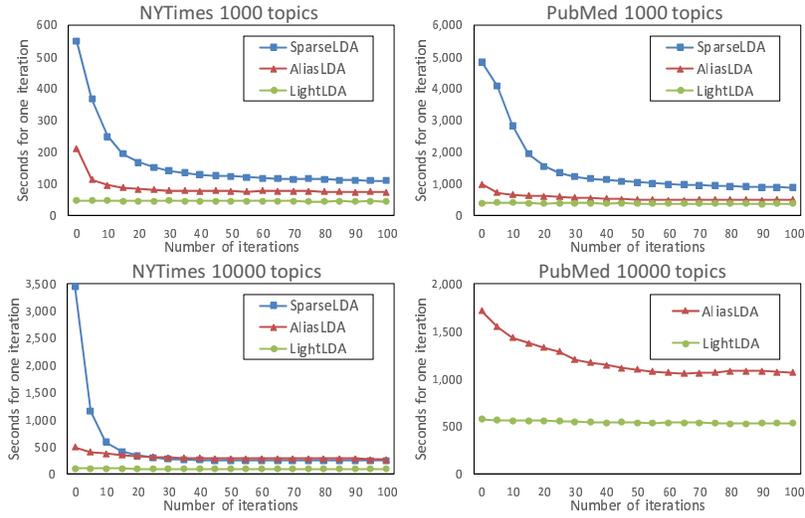

Figure 13: Running time for each of the first 100 iterations of SparseLDA, AliasLDA, LightLDA. The curve for SparseLDA was omitted from the $K = 10,000$-topic PubMed experiment, as it was too slow to show up in the plotted range.

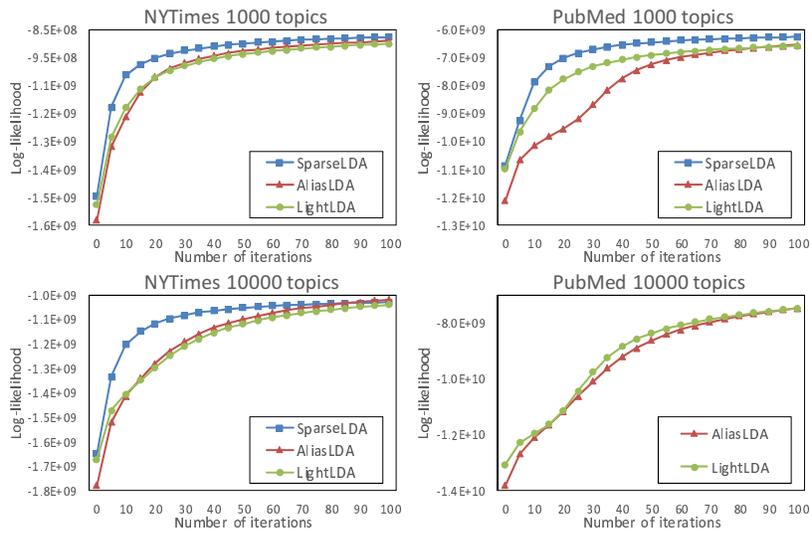

Figure 14: Log-likelihood versus iteration number for SparseLDA, AliasLDA, and LightLDA.